\documentclass[11pt, a4paper, logo, copyright, nonumbering]{deepseek}
\usepackage[numbers, square, sort&compress]{natbib}
\usepackage{dblfloatfix}
\usepackage{ulem}
\usepackage{caption}
\definecolor{citecolor}{HTML}{1976D2}
\hypersetup{
    colorlinks=true,
    linkcolor=black,
    filecolor=magenta,
    urlcolor=cyan,
    citecolor=citecolor,
}
\usepackage{dramatist}
\usepackage{xspace}
\usepackage{pifont}
\usepackage{multirow}
\usepackage{tcolorbox}
\usepackage{xltabular}
\usepackage{longtable}
\usepackage{hyperref}
\usepackage{diagbox}
\usepackage{makecell}
\usepackage{amsfonts}
\usepackage{amsmath}
\usepackage{amssymb}
\usepackage{lineno}

\usepackage[bottom]{footmisc}

\usepackage{CJKutf8}
\usepackage{subfigure}
\usepackage{setspace}
\usepackage{subcaption} 

\makeatletter
\def\@BTrule[#1]{%
  \ifx\longtable\undefined
    \let\@BTswitch\@BTnormal
  \else\ifx\hline\LT@hline
    \nobreak
    \let\@BTswitch\@BLTrule
  \else
     \let\@BTswitch\@BTnormal
  \fi\fi
  \global\@thisrulewidth=#1\relax
  \ifnum\@thisruleclass=\tw@\vskip\@aboverulesep\else
  \ifnum\@lastruleclass=\z@\vskip\@aboverulesep\else
  \ifnum\@lastruleclass=\@ne\vskip\doublerulesep\fi\fi\fi
  \@BTswitch}
\makeatother

\addto\extrasenglish{
}

 {\begin{list}{}%
         {\setlength{\leftmargin}{#1}}%
         \item[]%
 }
 {\end{list}}
 
\reportnumber{001} 

\renewcommand{\today}{}

\title{\centering DeepSeek-OCR: Contexts Optical Compression}

\author{Haoran Wei, Yaofeng Sun, Yukun Li\\
\small DeepSeek-AI\\
}

\renewcommand{\phi}{\varphi}

\renewcommand{\epsilon}{\varepsilon}
\renewcommand{\imath}{\mathrm{i}}

\newlength{\restsubwidth}
\newlength{\restsubheight}
\newlength{\restsubmoreheight}
\newcommand{\rest}[2]{%
        \settowidth{\restsubwidth}{\ensuremath{#2}}
        \settoheight{\restsubheight}{\ensuremath{{}_{#2}}}
        \ensuremath{{#1\hskip 0.5pt}_{\vrule\kern2pt\parbox[b][%
        4pt][b]{\the\restsubwidth}{%
                        \ensuremath{{}_{#2}}}}}
        }

\begin{abstract}
We present DeepSeek-OCR as an initial investigation into the feasibility of compressing long contexts via optical 2D mapping. DeepSeek-OCR consists of two components: DeepEncoder and DeepSeek3B-MoE-A570M as the decoder. Specifically, DeepEncoder serves as the core engine, designed to maintain low activations under high-resolution input while achieving high compression ratios to ensure an optimal and manageable  number of vision tokens. Experiments show that when the number of text tokens is within 10 times that of vision tokens (i.e., a compression ratio < 10$\times$),  the model can achieve decoding (OCR) precision of 97\%.  Even at a compression ratio of 20$\times$, the OCR accuracy still remains at about 60\%. This shows considerable promise for research areas such as historical long-context compression and memory forgetting mechanisms in LLMs. Beyond this, DeepSeek-OCR also demonstrates high practical value. On OmniDocBench, it surpasses GOT-OCR2.0 (256 tokens/page) using only 100 vision tokens, and outperforms MinerU2.0 (6000+ tokens per page on average) while utilizing fewer than 800 vision tokens. In production, DeepSeek-OCR can generate training data for LLMs/VLMs at a scale of 200k+ pages per day (a single A100-40G). Codes and model weights are publicly accessible at \url{http://github.com/deepseek-ai/DeepSeek-OCR}.
\end{abstract}

\begin{document}
\begin{CJK*}{UTF8}{gbsn}

\maketitle

\begin{figure}[h]
    \centering
    \subfigure[Compression on Fox benchmark]{
        \includegraphics[width=0.47\textwidth]{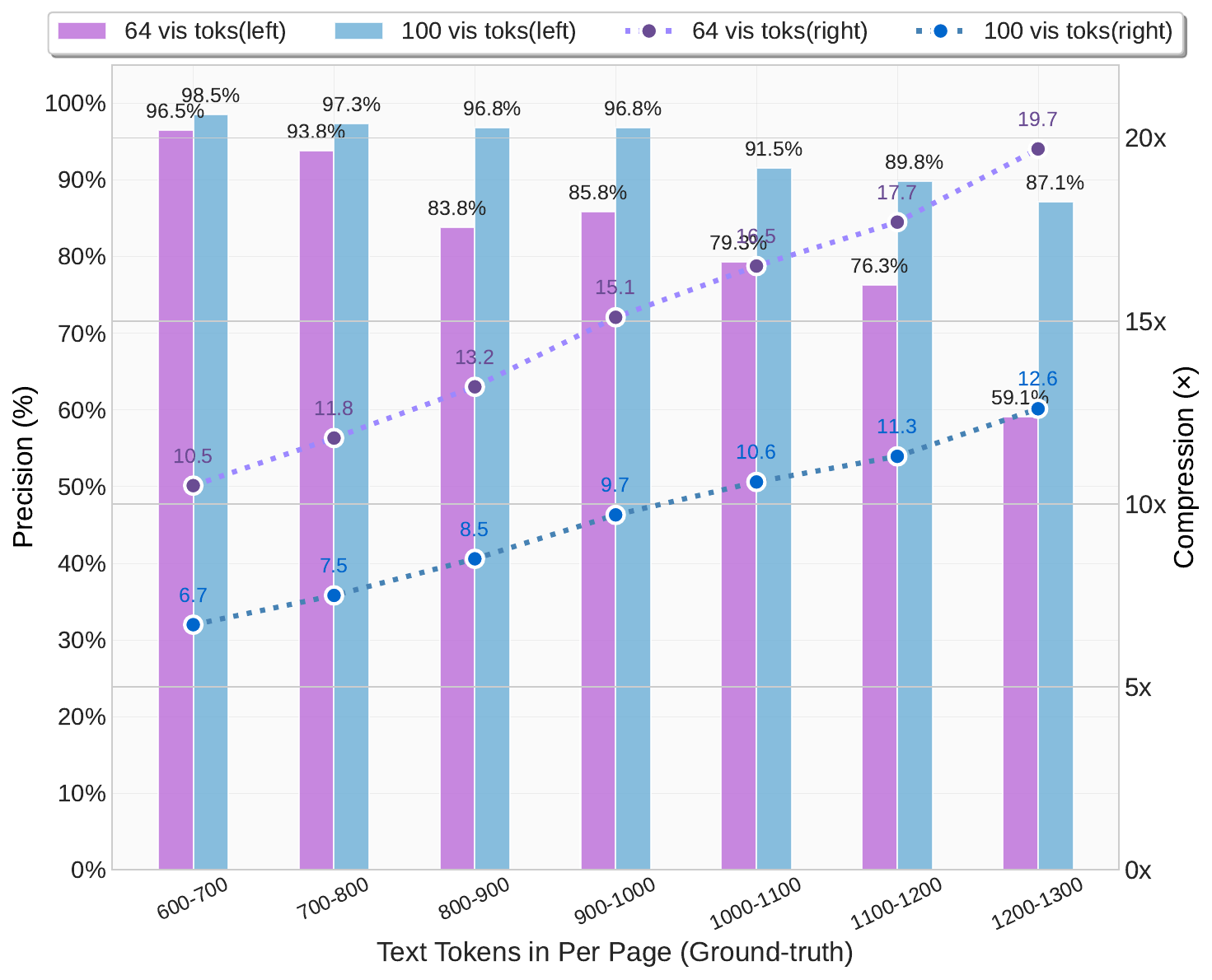}
        \label{fig:fox}
    }
    \subfigure[Performance on Omnidocbench]{
        \includegraphics[width=0.47\textwidth]{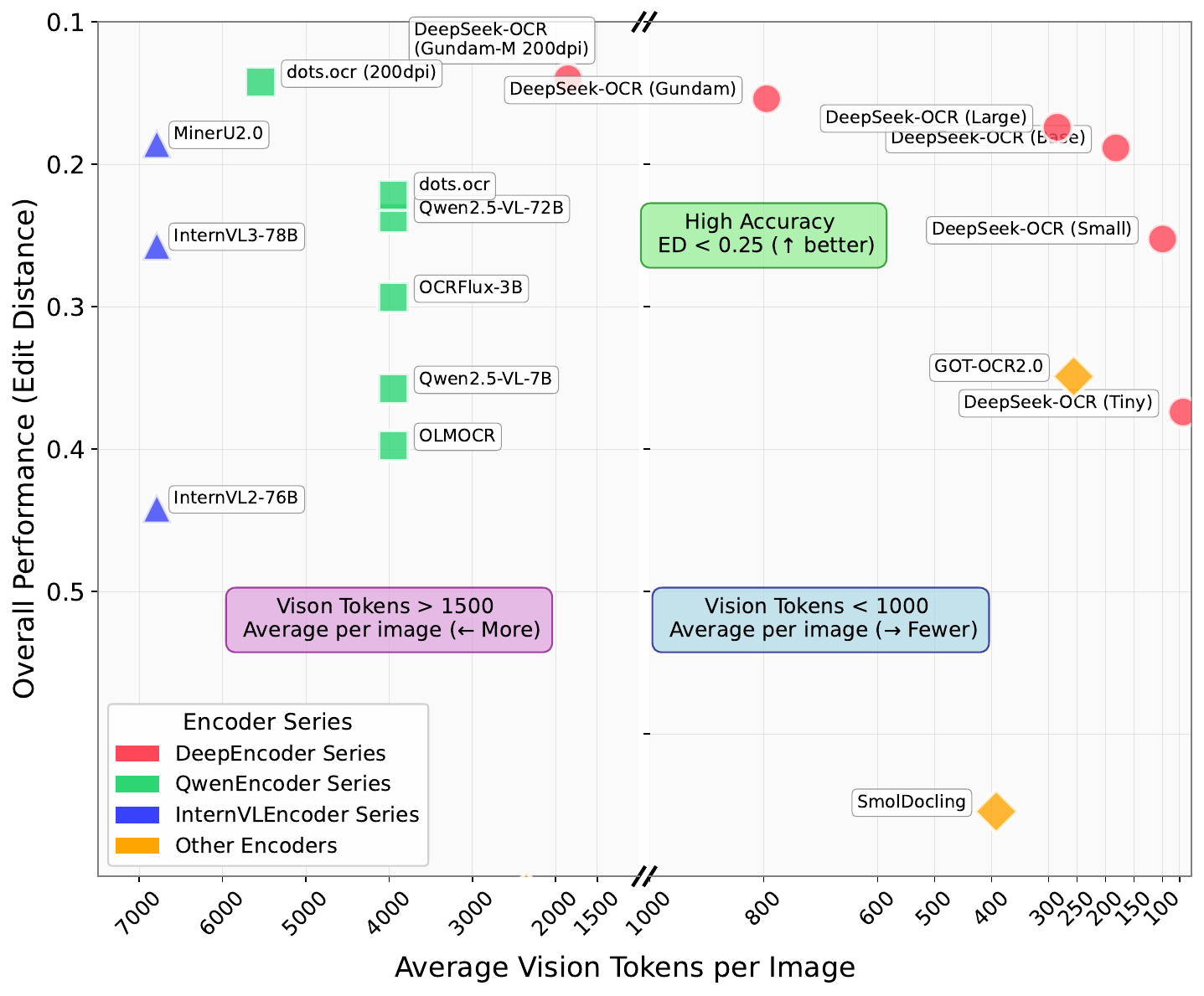}
        \label{fig:omni}
    }
    \caption{Figure (a) shows the compression ratio (number of text tokens in ground truth/number of vision tokens model used) testing on Fox~\cite{liu2024focus_fox} benchmark; Figure (b) shows performance comparisons on OmniDocBench~\cite{ouyang2025omnidocbench}. DeepSeek-OCR can achieve state-of-the-art performance among end-to-end models enjoying the fewest vision tokens.}
    \label{fig:1}
\end{figure}

\newpage

\begin{spacing}{0.9}
\tableofcontents
\end{spacing}

\newpage

\section{Introduction}

Current Large Language Models (LLMs) face significant computational challenges when processing long textual content due to quadratic scaling with sequence length. We explore a potential solution: leveraging visual modality as an efficient compression medium for textual information. A single image containing document text can represent rich information using substantially fewer tokens than the equivalent digital text, suggesting that optical compression through vision tokens could achieve much higher compression ratios.

This insight motivates us to reexamine vision-language models (VLMs) from an LLM-centric perspective, focusing on how vision encoders can enhance LLMs' efficiency in processing textual information rather than basic VQA~\cite{goyal2017making,masry2022chartqa,TextVQA,yu2023mm,kazemzadeh2014referitgame} what humans excel at. OCR tasks, as an intermediate modality bridging vision and language,  provide an ideal testbed for this vision-text compression paradigm, as they establish a natural compression-decompression mapping between visual and textual representations while offering quantitative evaluation metrics.

Accordingly, we present DeepSeek-OCR, a VLM designed as a preliminary proof-of-concept for efficient vision-text compression. Our work makes three primary contributions:

First, we provide comprehensive quantitative analysis of vision-text token compression ratios. Our method achieves 96\%+ OCR decoding precision at 9-10$\times$ text compression, $\sim$90\% at 10-12$\times$ compression, and $\sim$60\% at 20$\times$ compression on Fox~\cite{liu2024focus_fox} benchmarks featuring diverse document layouts (with actual accuracy being even higher when accounting for formatting differences between output and ground truth), as shown in Figure~\ref{fig:fox}. The results demonstrate that compact language models can effectively learn to decode compressed visual representations, suggesting that larger LLMs could readily acquire similar capabilities through appropriate pretraining design.

Second, we introduce DeepEncoder, a novel architecture that maintains low activation memory and minimal vision tokens even with high-resolution inputs. It serially connects window attention and global attention encoder components through a 16$\times$ convolutional compressor. This design ensures that the window attention component processes a large number of vision tokens, while the compressor reduces vision tokens before they enter the dense global attention component, achieving effective memory and token compression.

Third, we develop DeepSeek-OCR based on DeepEncoder and DeepSeek3B-MoE~\cite{liu2024deepseekv2,liu2024deepseekv3}. As shown in Figure~\ref{fig:omni}, it achieves state-of-the-art performance within end-to-end models on OmniDocBench while using the fewest vision tokens. Additionally, we equip the model with capabilities for parsing charts, chemical formulas, simple geometric figures, and natural images to enhance its practical utility further.  In production, DeepSeek-OCR can generate 33 million pages of data per day for LLMs or VLMs using 20 nodes (each with 8 A100-40G GPUs).

In summary, this work presents a preliminary exploration of using visual modality as an efficient compression medium for textual information processing in LLMs. Through DeepSeek-OCR, we demonstrate that vision-text compression can achieve significant token reduction (7-20$\times$) for different historical context stages, offering a promising direction for addressing long-context challenges in large language models. Our quantitative analysis provides empirical guidelines for VLM token allocation optimization, while the proposed DeepEncoder architecture showcases practical feasibility with real-world deployment capabilities. Although focused on OCR as a proof-of-concept, this paradigm opens new possibilities for rethinking how vision and language modalities can be synergistically combined to enhance computational efficiency in large-scale text processing and agent systems.

\begin{figure}[ht]
	\centering
    \includegraphics[width=1.0\linewidth]{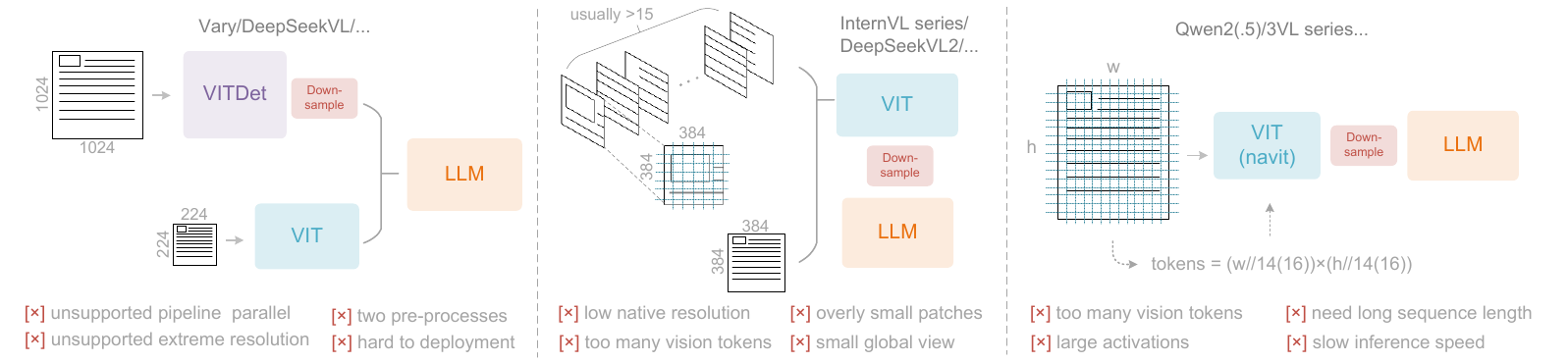}
	\caption{Typical vision encoders in popular VLMs. Here are three types of encoders commonly used in current open-source VLMs, all of which suffer from their respective deficiencies.}
	\label{fig:encoders}
\end{figure}

\section{Related Works}


\subsection{Typical Vision Encoders in VLMs}

\label{sec:related1}

Current open-source VLMs employ three main types of vision encoders, as illustrated in Figure~\ref{fig:encoders}. The first type is a dual-tower architecture represented by Vary~\cite{wei2024vary}, which utilizes parallel SAM~\cite{kirillov2023segment} encoder to increase visual vocabulary parameters for high-resolution image processing. While offering controllable parameters and activation memory, this approach suffers from significant drawbacks: it requires dual image preprocessing that complicates deployment and makes encoder pipeline parallelism challenging during training. The second type is tile-based method exemplified by InternVL2.0~\cite{chen2024internvl2}, which processes images by dividing them into small tiles for parallel computation, reducing activation memory under high-resolution settings. Although capable of handling extremely high resolutions, this approach has notable limitations due to its typically low native encoder resolution (below 512$\times$512), causing large images to be excessively fragmented and resulting in numerous vision tokens. The third type is adaptive resolution encoding represented by Qwen2-VL~\cite{wang2024qwen2}, which adopts the NaViT~\cite{dehghani2023patch} paradigm to directly process full images through patch-based segmentation without tile parallelization. While this encoder can handle diverse resolutions flexibly, it faces substantial challenges with large images due to massive activation memory consumption that can cause GPU memory overflow, and sequence packing requires extremely long sequence lengths during training. Long vision tokens will slow down both prefill and generation phases of inference.

\subsection{End-to-end OCR Models}
OCR, particularly document parsing task, has been a highly active topic in the image-to-text domain. With the advancement of VLMs, a large number of end-to-end OCR models have emerged, fundamentally transforming the traditional pipeline architecture (which required separate detection and recognition expert models) by simplifying OCR systems. Nougat~\cite{blecher2023nougat} first employs end-to-end framework for academic paper OCR on arXiv, demonstrating the potential of models in handling dense perception tasks. GOT-OCR2.0~\cite{wei2024general} expands the scope of OCR2.0 to include more synthetic image parsing tasks and designs an OCR model with performance-efficiency trade-offs, further highlighting the potential of end-to-end OCR researches. Additionally, general vision models such as Qwen-VL series~\cite{wang2024qwen2}, InternVL series~\cite{chen2024internvl2}, and many their derivatives continuously enhance their document OCR capabilities to explore dense visual perception boundaries. However, a crucial research question that current models have not addressed is: \textit{for a document containing 1000 words, how many vision tokens are at least needed for decoding?} This question holds significant importance for research in the principle that "\textit{a picture is worth a thousand words.}"

\begin{figure}[ht]
	\centering
    \includegraphics[width=1.0\linewidth]{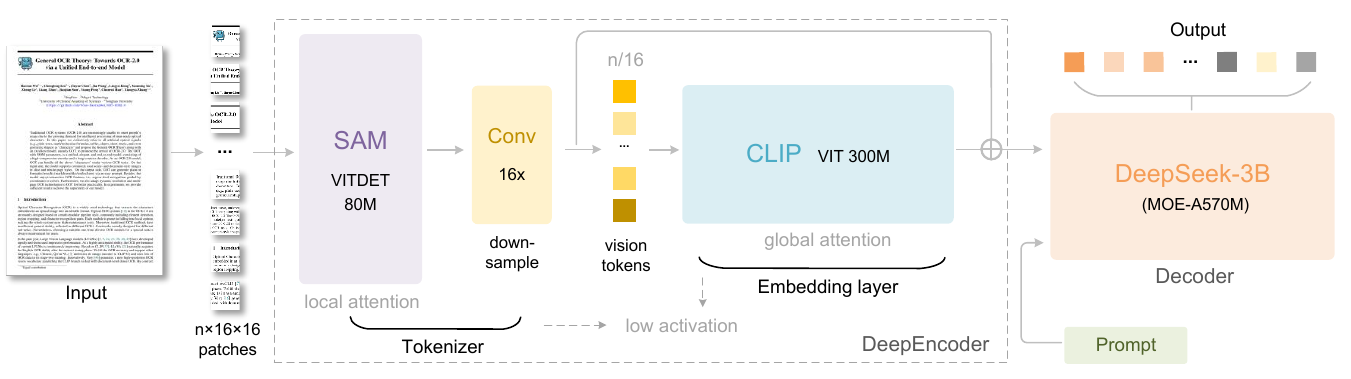}
	\caption{The architecture of DeepSeek-OCR. DeepSeek-OCR consists of a DeepEncoder and a DeepSeek-3B-MoE decoder. DeepEncoder is the core of DeepSeek-OCR, comprising three components: a SAM~\cite{kirillov2023segment} for perception dominated by window attention, a CLIP~\cite{radford2021learning} for knowledge  with dense global attention, and a 16$\times$ token compressor that bridges between them.}
	\label{fig:architecture}
\end{figure}

\section{Methodology}
\subsection{Architecture}
As shown in Figure~\ref{fig:architecture}, DeepSeek-OCR enjoys a unified end-to-end VLM architecture consisting of an encoder and a decoder. The encoder (namely DeepEncoder) is responsible for extracting image features and tokenizing as well as compressing visual representations. The decoder is used for generating the required result based on image tokens and prompts. DeepEncoder is approximately 380M in parameters, mainly composed of an 80M SAM-base~\cite{kirillov2023segment} and a 300M CLIP-large~\cite{radford2021learning} connected in series. The decoder adopts a 3B MoE~\cite{liu2024deepseekv2,liu2024deepseekv3} architecture with 570M activated parameters. In the following paragraphs, we will delve into the model components, data engineering, and training skills.

\subsection{DeepEncoder}
To explore the feasibility of contexts optical compression, we need a vision encoder with the following features: 1.Capable of processing high resolutions; 2.Low activation at high resolutions; 3.Few vision tokens; 4.Support for multiple resolution inputs; 5. Moderate parameter count. However, as described in the Section~\ref{sec:related1}, current open-source encoders cannot fully satisfy all these conditions. Therefore, we design a novel vision encoder ourselves, named DeepEncoder.

\begin{figure}[ht]
	\centering
    \includegraphics[width=1.0\linewidth]{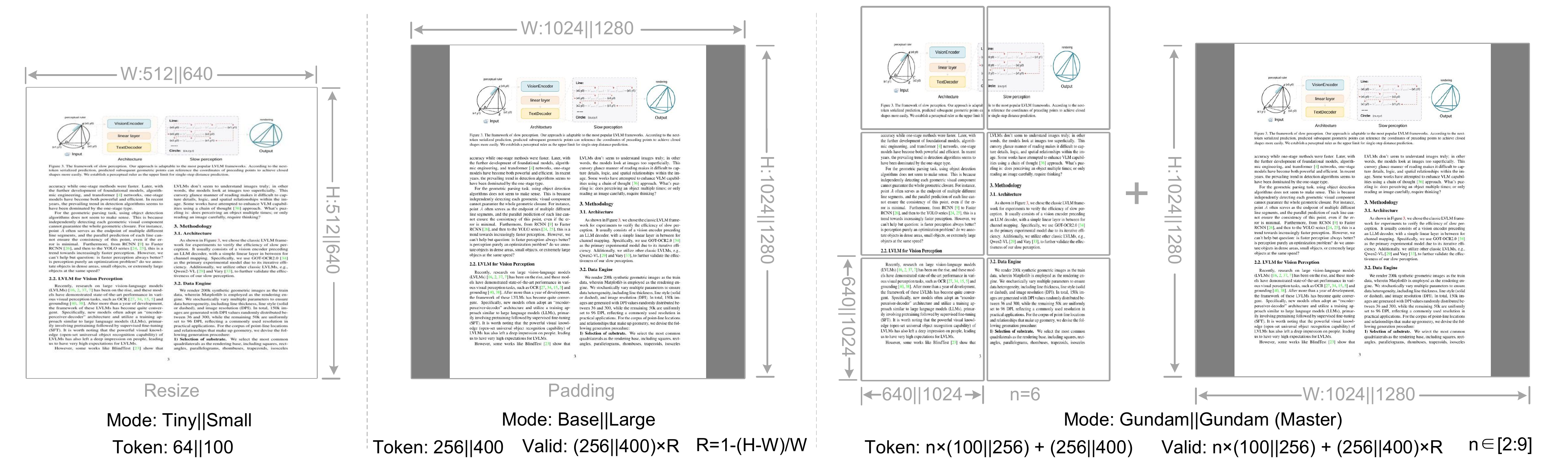}
	\caption{To test model performance under different compression ratios (requiring different numbers of vision tokens) and enhance the practicality of DeepSeek-OCR, we configure it with multiple resolution modes. }
	\label{fig:resolution}
\end{figure}

\subsubsection{Architecture of DeepEncoder}
DeepEncoder mainly consists of two components: a visual perception feature extraction component dominated by window attention, and a visual knowledge feature extraction component with dense global attention. To benefit from the pretraining gains of previous works, we use SAM-base (patch-size 16) and CLIP-large as the main architectures for the two components respectively. For CLIP, we remove the first patch embedding layer since its input is no longer images but output tokens from the previous pipeline. Between the two components, we borrow from Vary~\cite{wei2024vary} and use a 2-layer convolutional module to perform 16$\times$ downsampling of vision tokens. Each convolutional layer has a kernel size of 3, stride of 2, padding of 1, and channels increase from 256 to 1024. Assuming we input a 1024$\times$1024 image, the DeepEncoder will segment it into 1024/16$\times$1024/16=4096 patch tokens. Since the first half of encoder is dominated by window attention and only 80M, the activation is acceptable. Before entering global attention, the 4096 tokens go through the compression module and the token count becomes 4096/16=256, thus making the overall activation memory controllable.

\begin{table}[!h]\normalsize
	\centering	
        
 	\caption{Multi resolution support of DeepEncoder. For both research and application purposes, we design DeepEncoder with diverse native resolution and dynamic resolution modes.}
    \setlength{\abovecaptionskip}{0.2cm}
	\setlength{\tabcolsep}{1.0mm}{	
		
		\begin{tabular}{lcccc|cc}
			\toprule 
            \multirow{3}{*}{\textbf{Mode}} & \multicolumn{4}{c}{\textbf{Native Resolution}} &\multicolumn{2}{c}{\textbf{Dynamic Resolution}} \\
            \cmidrule(rl){2-5}  \cmidrule(rl){6-7}
             & Tiny & Small & Base & Large & Gundam & Gundam-M \\ 
			\midrule  
			Resolution& 512 & 640  & 1024 & 1280 & 640+1024 & 1024+1280 \\
			Tokens& 64 & 100 & 256 & 400 & n$\times$100+256& n$\times$256+400\\
			  Process& resize & resize & padding & padding & resize + padding & resize + padding\\

			\bottomrule		
	\end{tabular}}	

	\label{table1}
\end{table}

\subsubsection{Multiple resolution support}
Suppose we have an image with 1000 optical characters and we want to test how many vision tokens are needed for decoding. This requires the model to support a variable number of vision tokens. That is to say the DeepEncoder needs to support multiple resolutions. 


We meet the requirement aforementioned through dynamic interpolation of positional encodings, and design several resolution modes for simultaneous model training to achieve the capability of a single DeepSeek-OCR model supporting multiple resolutions. As shown in Figure~\ref{fig:resolution}, DeepEncoder mainly supports two major input modes: native resolution and dynamic resolution. Each of them contains multiple sub-modes.

Native resolution supports four sub-modes: Tiny, Small, Base, and Large, with corresponding resolutions and token counts of 512$\times$512 (64), 640$\times$640 (100), 1024$\times$1024 (256), and 1280$\times$1280 (400) respectively. Since Tiny and Small modes have relatively small resolutions, to avoid wasting vision tokens, images are processed by directly resizing the original shape. For Base and Large modes, in order to preserve the original image aspect ratio, images are padded to the corresponding size. After padding, the number of valid vision tokens is less than the actual number of vision tokens, with the calculation formula being:
\begin{equation}
N_{valid} = \lceil N_{actual} \times [1 - ((max(w, h) - min(w, h)) / (max(w, h)))] \rceil
\label{enq1}
\end{equation}
where $w$ and $h$ represent the width and height of the original input image.

Dynamic resolution can be composed of two native resolutions. For example, Gundam mode consists of n$\times$640$\times$640 tiles (local views) and a 1024$\times$1024 global view. The tiling method following InternVL2.0~\cite{chen2024internvl2}. Supporting dynamic resolution is mainly for application considerations, especially for ultra-high-resolution inputs (such as newspaper images). Tiling is a form of secondary window attention that can effectively reduce activation memory further. It's worth noting that due to our relatively large native resolutions, images won't be fragmented too much under dynamic resolution (the number of tiles is controlled within the range of 2 to 9). The vision token number output by the DeepEncoder under Gundam mode is: $n\times100+256$, where $n$ is the number of tiles. For images with both width and height smaller than 640, $n$ is set to 0, i.e., Gundam mode will degrade to Base mode. 

Gundam mode is trained together with the four native resolution modes to achieve the goal of one model supporting multiple resolutions. Note that Gundam-master mode (1024$\times$1024 local views+1280$\times$1280 global view) is obtained through continued training on a trained DeepSeek-OCR model. This is mainly for load balancing, as Gundam-master's resolution is too large and training it together would slow down the overall training speed.

\subsection{The MoE Decoder}

Our decoder uses the DeepSeekMoE~\cite{liu2024deepseekv2,liu2024deepseekv3}, specifically DeepSeek-3B-MoE. During inference, the model activates 6 out of 64 routed experts and 2 shared experts, with about 570M activated parameters. The 3B DeepSeekMoE is very suitable for domain-centric (OCR for us) VLM research, as it obtains the expressive capability of a 3B model while enjoying the inference efficiency of a 500M small model.

The decoder reconstructs the original text representation from the compressed latent vision tokens of DeepEncoder as:
\begin{equation}
f_{\text{dec}}: \mathbb{R}^{n \times d_{\text{latent}}} \rightarrow \mathbb{R}^{N \times d_{\text{text}}}; \quad \hat{\mathbf{X}} = f_{\text{dec}}(\mathbf{Z}) \quad \text{where } n \le N
\label{enq2}
\end{equation}
where $\mathbf{Z} \in \mathbb{R}^{n \times d_{\text{latent}}}$ are the compressed latent(vision) tokens from DeepEncoder and $\hat{\mathbf{X}} \in \mathbb{R}^{N \times d_{\text{text}}}$ is the reconstructed text representation. The function $f_{\text{dec}}$ represents a non-linear mapping that can be effectively learned by compact language models through OCR-style training. It is reasonable to conjecture that LLMs, through specialized pretraining optimization, would demonstrate more natural integration of such capabilities.

\subsection{Data Engine}
\label{data}
We constructe complex and diverse training data for DeepSeek-OCR, including OCR 1.0 data, which mainly consists of traditional OCR tasks such as scene image OCR and document OCR; OCR 2.0 data, which mainly includes parsing tasks for complex artificial images, such as common charts, chemical formulas, and plane geometry parsing data; General vision data, which is mainly used to inject certain general image understanding capabilities into DeepSeek-OCR and preserve the general vision interface.

\begin{figure}[h]
    \centering
    \subfigure[Ground truth image]{
        \includegraphics[width=0.47\textwidth]{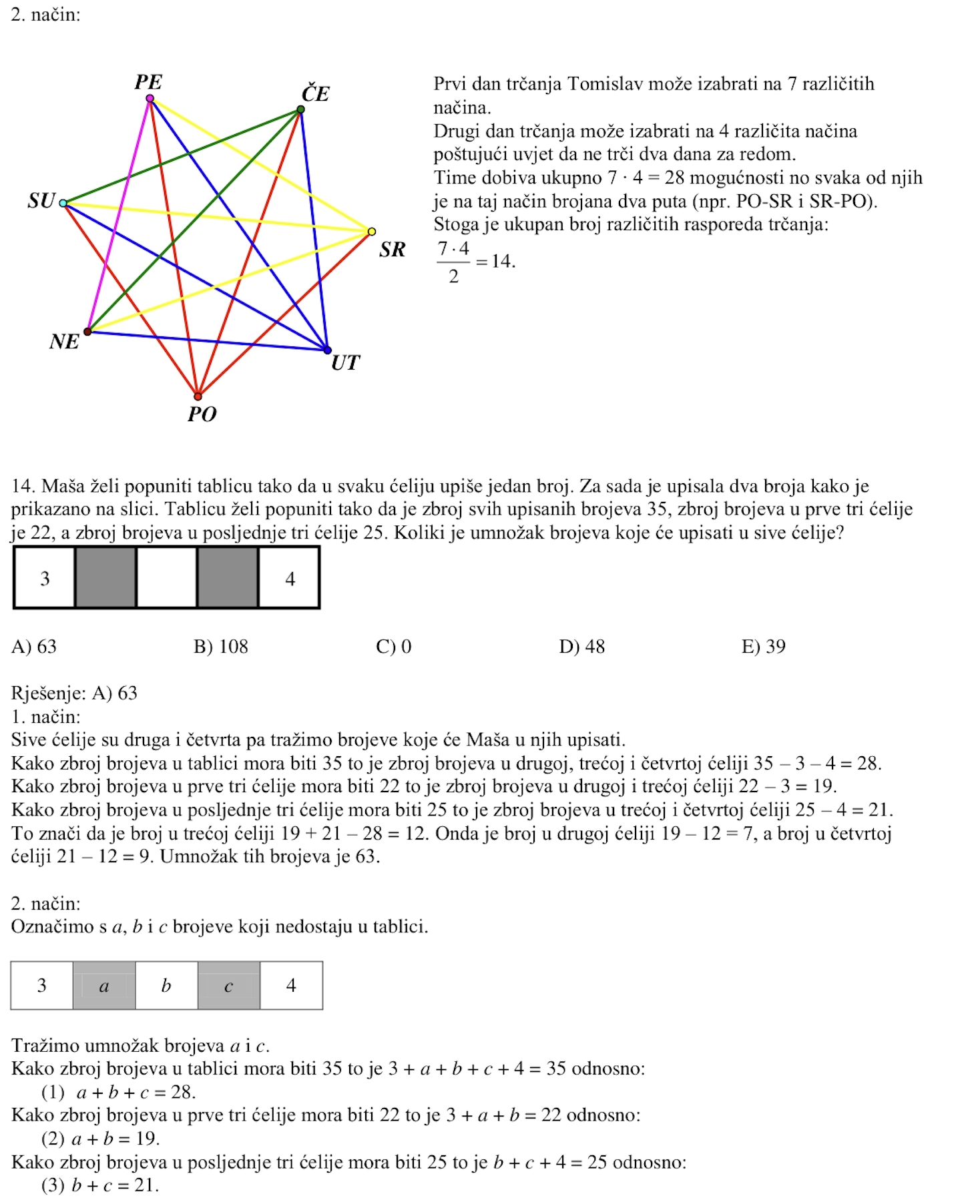}
        \label{fig:demo1}
    }
    \subfigure[Fine annotations with layouts]{
        \includegraphics[width=0.47\textwidth]{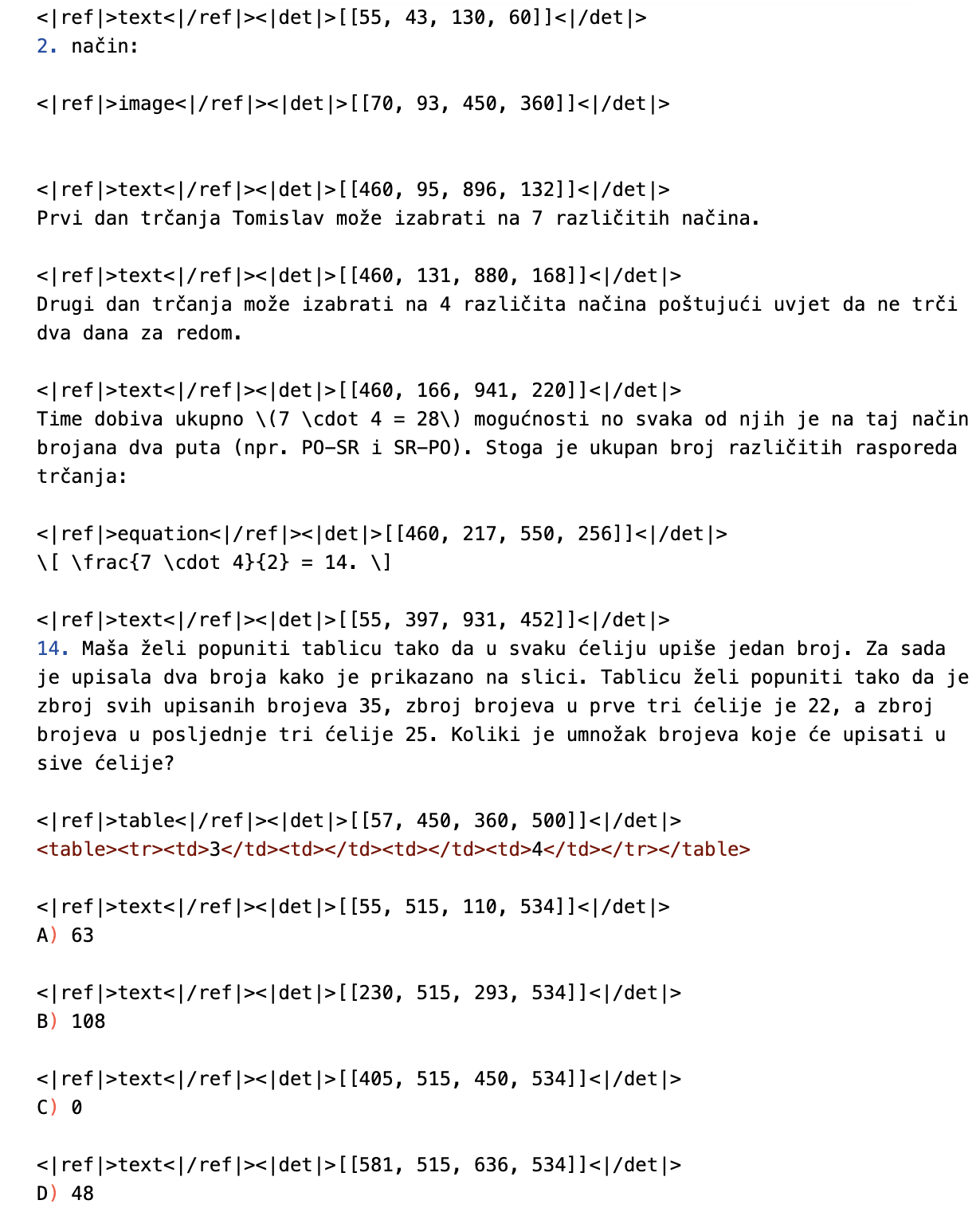}
        \label{fig:demo2}
    }
    \caption{OCR 1.0 fine annotations display. We format the ground truth into an interleaved layout and text format, where each paragraph of text is preceded by the coordinates and label of it in the original image. All coordinates are normalized into 1000 bins.}
    \label{fig:demo}

\end{figure}

\subsubsection{OCR 1.0 data}
Document data is the top priority for DeepSeek-OCR. We collect 30M pages of diverse PDF data covering about 100 languages from the Internet, with Chinese and English accounting for approximately 25M and other languages accounting for 5M. For this data, we create two types of ground truth: coarse annotations and fine annotations. Coarse annotations are extracted directly from the full dataset using \textit{fitz}, aimed at teaching the model to recognize optical text, especially in minority languages. Fine annotations include 2M pages each for Chinese and English, labeled using advanced layout models (such as PP-DocLayout~\cite{sun2025pp}) and OCR models (such as MinuerU~\cite{wang2024mineru} and GOT-OCR2.0~\cite{wei2024general}) to construct detection and recognition interleaved data. For minority languages, in the detection part, we find that the layout model enjoys certain generalization capabilities. In the recognition part, we use \textit{fitz} to create small patch data to train a GOT-OCR2.0, then use the trained model to label small patches after layout processing, employing a model flywheel to create 600K data samples. During the training of DeepSeek-OCR, coarse labels and fine labels are distinguished using different prompts. The ground truth for fine annotation image-text pairs can be seen in Figure~\ref{fig:demo}. We also collect 3M \textit{Word} data, constructing high-quality image-text pairs without layout by directly extracting content. This data mainly brings benefits to formulas and HTML-formatted tables. Additionally, we select some open-source data~\cite{poznanski2025olmocr,wei2024small} as supplements. 

For natural scene OCR, our model mainly supports Chinese and English. The image data sources come from LAION~\cite{schuhmann2021laion} and Wukong~\cite{gu2022wukong}, labeled using PaddleOCR~\cite{cui2025paddleocr}, with 10M data samples each for Chinese and English. Like document OCR, natural scene OCR can also control whether to output detection boxes through prompts.

\subsubsection{OCR 2.0 data}

\begin{figure}[!t]
    \centering
    \subfigure[Image-text ground truth of chart]{
        \includegraphics[width=0.48\textwidth]{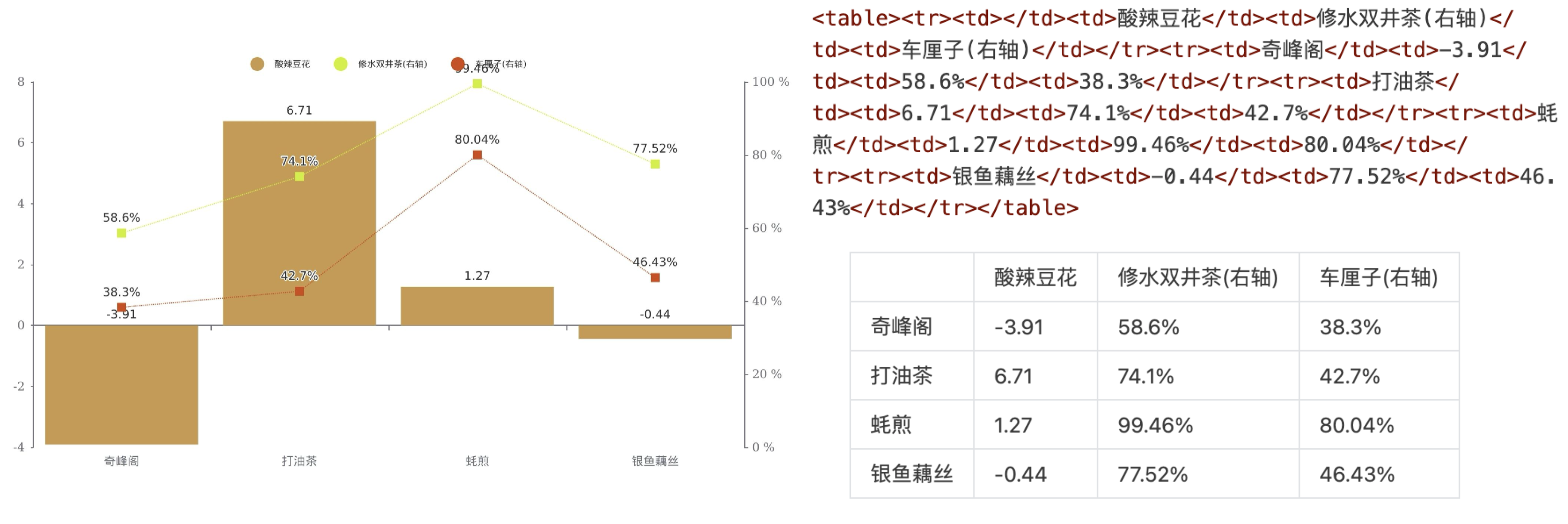}
        \label{fig:demo2-1}
    }
    \subfigure[Image-text ground truth of geometry]{
        \includegraphics[width=0.48\textwidth]{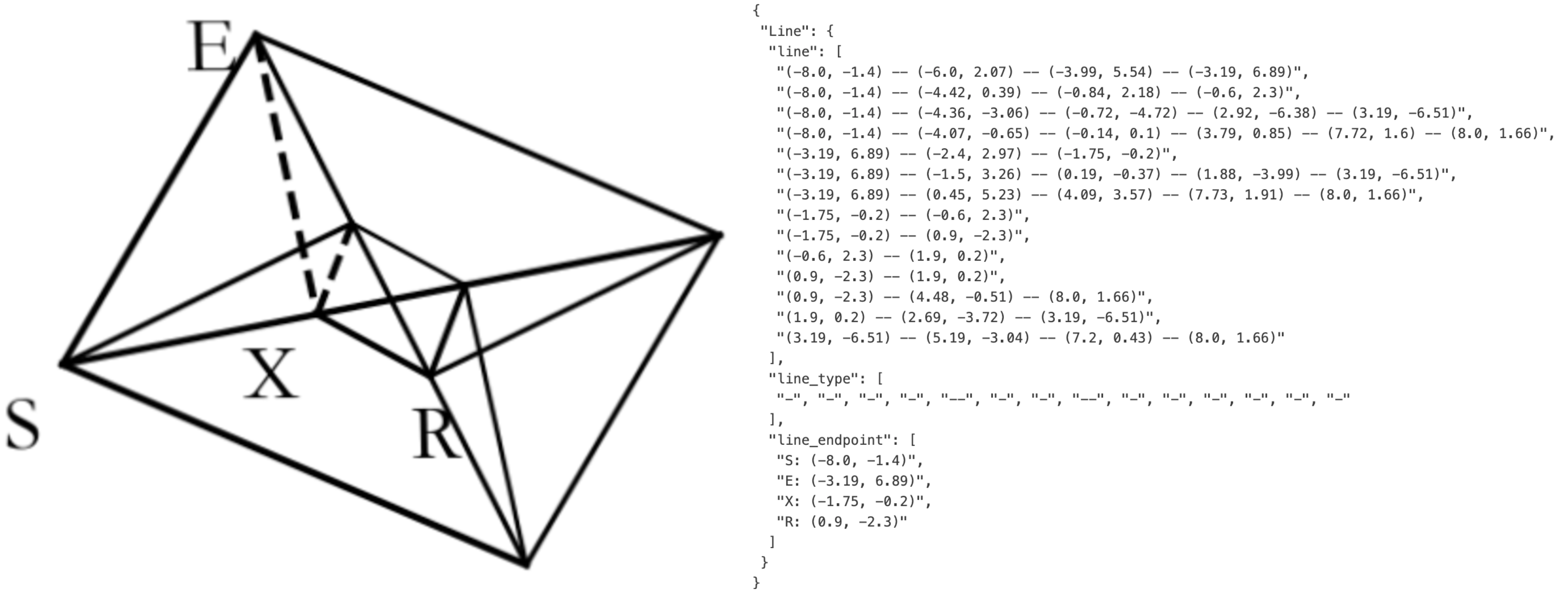}
        \label{fig:demo2-2}
    }
    \caption{For charts, we do not use OneChart's~\cite{chen2024onechart} dictionary format, but instead use HTML table format as labels, which can save a certain amount of tokens. For plane geometry, we convert the ground truth to dictionary format, where the dictionary contains keys such as line segments, endpoint coordinates, line segment types, etc., for better readability. Each line segment is encoded using the Slow Perception~\cite{wei2024slow} manner.}
    \label{fig:demo2}
\end{figure}

Following GOT-OCR2.0~\cite{wei2024general}, we refer to chart, chemical formula, and plane geometry parsing data as OCR 2.0 data. For chart data, following OneChart~\cite{chen2024onechart}, we use pyecharts and matplotlib to render 10M images, mainly including commonly used line, bar, pie, and composite charts. We define chart parsing as image-to-HTML-table conversion task, as shown in Figure~\ref{fig:demo2-1}. For chemical formulas, we utilize SMILES format from PubChem as the data source and render them into images using RDKit, constructing 5M image-text pairs. For plane geometry images, we follow Slow Perception~\cite{wei2024slow} for generation. Specifically, we use perception-ruler size as 4 to model each line segment. To increase the diversity of rendered data, we introduce geometric translation-invariant data augmentation, where the same geometric image is translated in the original image, corresponding to the same ground truth drawn at the centered position in the coordinate system. Based on this, we construct a total of 1M plane geometry parsing data, as illustrated in Figure~\ref{fig:demo2-2}.

\subsubsection{General vision data}
DeepEncoder can benefit from CLIP's pretraining gains and has sufficient parameters to incorporate general visual knowledge. Therefore, we also prepare some corresponding data for DeepSeek-OCR. Following DeepSeek-VL2~\cite{wu2024deepseek}, we generate relevant data for tasks such as caption, detection, and grounding. Note that DeepSeek-OCR is not a general VLM model, and this portion of data accounts for only 20\% of the total data. We introduce such type of data mainly to preserve the general vision interface, so that researchers interested in our model and general vision task can conveniently advance their work in the future.

\subsubsection{Text-only data}
To ensure the model's language capabilities, we introduced 10\% of in-house text-only pretrain data, with all data processed to a length of 8192 tokens, which is also the sequence length for DeepSeek-OCR. In summary, when training DeepSeek-OCR, OCR data accounts for 70\%, general vision data accounts for 20\%, and text-only data accounts for 10\%.

\subsection{Training Pipelines}
Our training pipeline is very simple and consists mainly of two stages: a).Training DeepEncoder independently; b).Training the DeepSeek-OCR. Note that the Gundam-master mode is obtained by continuing training on a pre-trained DeepSeek-OCR model with 6M sampled data. Since the training protocol is identical to other modes, we omit the detailed description hereafter.

\subsubsection{Training DeepEncoder}
Following Vary~\cite{wei2024vary}, we utilize a compact language model~\cite{OPT-IML} and use the next token prediction framework to train DeepEncoder.  In this stage, we use all OCR 1.0 and 2.0 data aforementioned, as well as 100M general data sampled from the LAION~\cite{schuhmann2021laion} dataset. All data is trained for 2 epochs with a batch size of 1280, using the AdamW~\cite{AdamW} optimizer with cosine annealing scheduler~\cite{loshchilov2016sgdr} and a learning rate of 5e-5. The training sequence length is 4096.

\subsubsection{Training DeepSeek-OCR}
After DeepEncoder is ready, we use data mentioned in Section~\ref{data} to train the DeepSeek-OCR. with the entire training process conducted on the HAI-LLM~\cite{highflyer2023haillm} platform. The entire model uses pipeline parallelism (PP) and is divided into 4 parts, with DeepEncoder taking two parts and the decoder taking two parts. For DeepEncoder, we treat SAM and the compressor as the vision tokenizer, place them in PP0 and freeze their parameters, while treating the CLIP part as input embedding layer and place it in PP1 with unfrozen weights for training. For the language model part, since DeepSeek3B-MoE has 12 layers, we place 6 layers each on PP2 and PP3. We use 20 nodes (each with 8 A100-40G GPUs) for training, with a data parallelism (DP) of 40 and a global batch size of 640. We use the AdamW optimizer with a step-based scheduler and an initial learning rate of 3e-5. For text-only data, the training speed is 90B tokens/day, while for multimodal data, the training speed is 70B tokens/day.

\begin{table}[!h]\normalsize
	\centering	
        
 	\caption{We test DeepSeek-OCR's vision-text compression ratio using all English documents with 600-1300 tokens from the Fox~\cite{liu2024focus_fox} benchmarks. Text tokens represent the number of tokens after tokenizing the ground truth text using DeepSeek-OCR's tokenizer. Vision Tokens=64 or 100 respectively represent the number of vision tokens output by DeepEncoder after resizing input images to 512$\times$512 and 640$\times$640.}
    \setlength{\abovecaptionskip}{0.2cm}
	\setlength{\tabcolsep}{1.0mm}{	
		
		\begin{tabular}{ccc|ccc}
			\toprule 
            \multirow{3}{*}{\textbf{Text Tokens}} & \multicolumn{2}{c}{\textbf{Vision Tokens =64}} &\multicolumn{2}{c}{\textbf{Vision Tokens=100}} \\
            \cmidrule(rl){2-3}  \cmidrule(rl){4-5}
             & Precision & Compression & Precision & Compression & Pages \\ 
			\midrule  
			600-700& 96.5\% & 10.5$\times$  & 98.5\% & 6.7$\times$ & 7 \\
			700-800& 93.8\%  & 11.8$\times$ & 97.3\% & 7.5$\times$ & 28\\
			  800-900& 83.8\% & 13.2$\times$ & 96.8\% & 8.5$\times$ & 28 \\
            900-1000& 85.9\% & 15.1$\times$ & 96.8\% & 9.7$\times$ & 14 \\
            1000-1100& 79.3\% & 16.5$\times$ & 91.5\% & 10.6$\times$ & 11 \\
            1100-1200& 76.4\% & 17.7$\times$ & 89.8\% & 11.3$\times$ & 8 \\
            1200-1300& 59.1\% & 19.7$\times$ & 87.1\% & 12.6$\times$ & 4 \\

			\bottomrule		
	\end{tabular}}		

	\label{table2}
\end{table}

\begin{table}[!t]
    \small
	\centering	
        
 	\caption{We use OmniDocBench~\cite{ouyang2025omnidocbench} to test the performance of DeepSeek-OCR on real document parsing tasks. All metrics in the table are edit distances, where smaller values indicate better performance. "Tokens" represents the average number of vision tokens used per page, and "$^{\dagger{\mathrm{200dpi}}}$" means using \textit{fitz} to interpolate the original image to 200dpi. For the DeepSeek-OCR model, the values in parentheses in the "Tokens" column represent valid vision tokens, calculated according to Equation~\ref{enq1}.}
    \setlength{\abovecaptionskip}{0.2cm}
	\setlength{\tabcolsep}{0.5mm}{	
		
		\begin{tabular}{lccccccccccc}
			\toprule 
            \multirow{3}{*}{\textbf{Model}} & \multirow{3}{*}{\textbf{Tokens}} & \multicolumn{5}{c}{\textbf{English}} &\multicolumn{5}{c}{\textbf{Chinese}} \\
            \cmidrule(rl){3-7}  \cmidrule(rl){8-12} 
            & & \cellcolor{yellow!30} overall  & text & formula & table & order & \cellcolor{yellow!30} overall & text & formula & table &order \\
			\midrule  
            \multicolumn{12}{c}{\textbf{Pipline Models}} \\ 
            \midrule 
			Dolphin~\cite{feng2025dolphin}& - & 0.356  & 0.352 & 0.465 & 0.258 &0.35 & 0.44 & 0.44 & 0.604 & 0.367 & 0.351 \\
			Marker~\cite{marker}& -  & 0.296 & 0.085 & 0.374 & 0.609 & 0.116 & 0.497 & 0.293 & 0.688 & 0.678 & 0.329\\
			  Mathpix~\cite{mathpix}& - & 0.191 & 0.105 & 0.306 & 0.243 & 0.108 & 0.364 & 0.381 & 0.454 & 0.32 & 0.30 \\
            MinerU-2.1.1~\cite{wang2024mineru}& - & 0.162 & 0.072 & 0.313 & 0.166 & 0.097 & 0.244 & 0.111 & 0.581 & 0.15 & 0.136 \\
            MonkeyOCR-1.2B~\cite{li2025monkeyocr}& - & 0.154 & 0.062 & 0.295 & 0.164 & 0.094 & 0.263 & 0.179 & 0.464 & 0.168 & 0.243 \\
            PPstructure-v3~\cite{cui2025paddleocr}& - & 0.152 & 0.073 & 0.295 & 0.162 & 0.077 & 0.223 & 0.136 & 0.535 & 0.111 & 0.11 \\
			\midrule  
            \multicolumn{12}{c}{\textbf{End-to-end Models}} \\ 
            \midrule 
            Nougat~\cite{blecher2023nougat}& 2352 & 0.452 & 0.365 & 0.488 & 0.572 & 0.382 & 0.973 & 0.998 & 0.941 & 1.00 & 0.954\\
            SmolDocling~\cite{nassar2025smoldocling}& 392 & 0.493 & 0.262 & 0.753 & 0.729 & 0.227 & 0.816 & 0.838 & 0.997 & 0.907 & 0.522 \\
            InternVL2-76B~\cite{chen2024internvl2}& 6790 & 0.44 & 0.353 & 0.543 &  0.547 & 0.317 & 0.443 & 0.29 & 0.701 & 0.555 & 0.228  \\
            Qwen2.5-VL-7B~\cite{Qwen2.5-VL}& 3949 & 0.316 & 0.151 & 0.376 & 0.598 & 0.138 & 0.399 & 0.243 & 0.5 & 0.627 & 0.226 \\
            OLMOCR~\cite{poznanski2025olmocr}& 3949 & 0.326 & 0.097 & 0.455 & 0.608 & 0.145 & 0.469 & 0.293 & 0.655 & 0.652 & 0.277 \\
            GOT-OCR2.0~\cite{wei2024general}& 256 & 0.287 & 0.189 & 0.360 & 0.459 & 0.141 & 0.411 & 0.315 & 0.528 & 0.52 & 0.28 \\
            OCRFlux-3B~\cite{ocrflux}& 3949 & 0.238 & 0.112 & 0.447 & 0.269 &0.126 & 0.349 & 0.256 & 0.716 & 0.162 & 0.263 \\
            GPT4o~\cite{GPT4}& - & 0.233 & 0.144 & 0.425 & 0.234 & 0.128 & 0.399 & 0.409 & 0.606 & 0.329 & 0.251 \\
            InternVL3-78B~\cite{zhu2025internvl3}& 6790 & 0.218 & 0.117 & 0.38 & 0.279 & 0.095 & 0.296 & 0.21 & 0.533 & 0.282 & 0.161 \\
            Qwen2.5-VL-72B~\cite{Qwen2.5-VL}& 3949 & 0.214 & 0.092 & 0.315 & 0.341 & 0.106 & 0.261 & 0.18 & 0.434 & 0.262 & 0.168 \\
            dots.ocr~\cite{dots}& 3949 & 0.182 & 0.137 & 0.320 & 0.166 & 0.182 & 0.261 & 0.229 & 0.468 & 0.160 & 0.261 \\
            Gemini2.5-Pro~\cite{google_gemini_web}& - & 0.148 & 0.055  & 0.356 & 0.13 & 0.049 & 0.212 & 0.168 & 0.439 & 0.119 & 0.121 \\
            MinerU2.0~\cite{wang2024mineru}& 6790 & 0.133 & 0.045 & 0.273 & 0.15 & 0.066 & 0.238 & 0.115 & 0.506 & 0.209 & 0.122 \\
            dots.ocr{\color{black}$^{\dagger{\mathrm{200dpi}}}$}~\cite{dots}& 5545 & 0.125 & \bf{0.032} & 0.329 & \bf{0.099} & \bf{0.04} & 0.16 & \bf{0.066} & 0.416 & 0.092 & \bf{0.067} \\
            \midrule 
            \rowcolor{blue!4}
            \multicolumn{12}{c}{\textbf{DeepSeek-OCR (end2end)}} \\ 
            \midrule 
            Tiny& \bf{64} & 0.386 & 0.373 & 0.469 & 0.422 & 0.283 & 0.361 & 0.307 & 0.635 & 0.266 & 0.236 \\
            Small& 100 & 0.221 &0.142 & 0.373 & 0.242 & 0.125 & 0.284 & 0.24 & 0.53 & 0.159 & 0.205\\
            Base& 256(182) & 0.137 & 0.054 & 0.267 & 0.163 & 0.064 & 0.24 & 0.205 & 0.474 & 0.1 & 0.181 \\
            Large& 400(285) & 0.138 & 0.054 & 0.277 & 0.152 & 0.067 & 0.208 & 0.143 & 0.461 & 0.104 & 0.123 \\
            Gundam& 795 & 0.127 & 0.043 & 0.269 & 0.134 & 0.062 & 0.181 & 0.097 & 0.432 & 0.089 & 0.103 \\
            Gundam-M{\color{black}$^{\dagger{\mathrm{200dpi}}}$}& 1853 & \bf{0.123} & 0.049 & \bf{0.242} & 0.147 &0.056 & \bf{0.157} & 0.087& \bf{0.377} & \bf{0.08} & 0.085 \\
            

			\bottomrule		
	\end{tabular}}		

	\label{table3}
\end{table}

\section{Evaluation}

\subsection{Vision-text Compression Study}
\label{vision-text compression}
We select Fox~\cite{liu2024focus_fox} benchmarks to verify DeepSeek-OCR's compression-decompression capability for text-rich documents, in order to preliminarily explore the  feasibility and boundaries of contexts optical compression. We use the English document portion of Fox, tokenize the ground truth text with DeepSeek-OCR's tokenizer (vocabulary size of approximately 129k), and select documents with 600-1300 tokens for testing, which happens to be 100 pages. Since the number of text tokens is not large, we only need to test performance in Tiny and Small modes, where Tiny mode corresponds to 64 tokens and Small mode corresponds to 100 tokens. We use the prompt without layout: "<image>\verb|\n|Free OCR." to control the model's output format. Nevertheless, the output format still cannot completely match Fox benchmarks, so the actual performance would be somewhat higher than the test results.

As shown in Table~\ref{table2}, within a 10$\times$ compression ratio, the model's decoding precision can reach approximately 97\%, which is a very promising result. In the future, it may be possible to achieve nearly 10$\times$ lossless contexts compression through text-to-image approaches. When the compression ratio exceeds 10$\times$, performance begins to decline, which may have two reasons: one is that the layout of long documents becomes more complex, and another reason may be that long texts become blurred at 512$\times$512 or 640$\times$640 resolution. The first issue can be solved by rendering texts onto a single layout page, while we believe the second issue will become a feature of the forgetting mechanism. When compressing tokens by nearly 20$\times$, we find that precision can still approach 60\%. These results indicate that optical contexts compression is a very promising and worthwhile research direction, and this approach does not bring any overhead because it can leverage VLM infrastructure, as multimodal systems inherently require an additional vision encoder.

\begin{table}[!h]\small
	\centering	
        
 	\caption{Edit distances for different categories of documents in OmniDocBench. The results show that some types of documents can achieve good performance with just 64 or 100 vision tokens, while others require Gundam mode.}
    \setlength{\abovecaptionskip}{0.2cm}
	\setlength{\tabcolsep}{0.5mm}{	
		
            \begin{tabular}{c|*{10}{c}}  %
            \toprule 
            \diagbox{Mode}{Type} & Book & Slides & \makecell{Financial \\ Report} & Textbook & \makecell{Exam \\ Paper} & Magazine & \makecell{Academic \\ Papers} & Notes & Newspaper & Overall \\ 
            \midrule
			Tiny& 0.147 & 0.116  & 0.207 & 0.173 & 0.294 & 0.201 & 0.395 & 0.297 & 0.94 & 0.32 \\
			Small& 0.085  & 0.111 & 0.079 & 0.147 & 0.171 & 0.107 & 0.131 & 0.187 & 0.744 & 0.205\\
			  Base& 0.037 & 0.08 & 0.027 & 0.1 & 0.13 & 0.073 & 0.052 & 0.176 & 0.645 & 0.156 \\
            Large& 0.038 & 0.108 & 0.022 &0.084 & 0.109& 0.06 & 0.053 & 0.155 & 0.353 & 0.117 \\
            Gundam& 0.035 & 0.085 & 0.289 & 0.095 & 0.094 & 0.059 & 0.039 & 0.153 & 0.122 & 0.083 \\
            Guandam-M& 0.052 & 0.09 & 0.034 &0.091 & 0.079 & 0.079 & 0.048 & 0.1 & 0.099 & 0.077 \\


			\bottomrule		
	\end{tabular}}		

	\label{table4}
\end{table}

\subsection{OCR Practical Performance}

DeepSeek-OCR is not only an experimental model; it has strong practical capabilities and can construct data for LLM/VLM pretraining. To quantify OCR performance, we test DeepSeek-OCR on OmniDocBench~\cite{ouyang2025omnidocbench}, with results shown in Table~\ref{table3}. Requiring only 100 vision tokens (640$\times$640 resolution), DeepSeek-OCR surpasses GOT-OCR2.0~\cite{wei2024general} which uses 256 tokens; with 400 tokens (285 valid tokens, 1280$\times$1280 resolution), it achieves on-par performance with state-of-the-arts on this benchmark. Using fewer than 800 tokens (Gundam mode), DeepSeek-OCR outperforms MinerU2.0~\cite{wang2024mineru} which needs nearly 7,000 vision tokens. These results demonstrate that our DeepSeek-OCR model is powerful in practical applications, and because the higher tokens compression, it enjoys a higher research ceiling. 

As shown in Table~\ref{table4}, some categories of documents require very few tokens to achieve satisfactory performance, such as slides which only need 64 vision tokens. For book and report documents, DeepSeek-OCR can achieve good performance with only 100 vision tokens. Combined with the analysis from Section~\ref{vision-text compression}, this may be because most text tokens in these document categories are within 1,000, meaning the vision-token compression ratio does not exceed 10$\times$. For newspapers, Gundam or even Gundam-master mode is required to achieve acceptable edit distances, because the text tokens in newspapers are 4-5,000, far exceeding the 10$\times$ compression of other modes. These experimental results further demonstrate the boundaries of contexts optical compression, which may provide effective references for researches on the vision token optimization in VLMs and context compression, forgetting mechanisms in LLMs.

\subsection{Qualitative Study}


\subsubsection{Deep parsing}
\begin{figure}[p]
	\centering
    \includegraphics[width=1.0\linewidth]{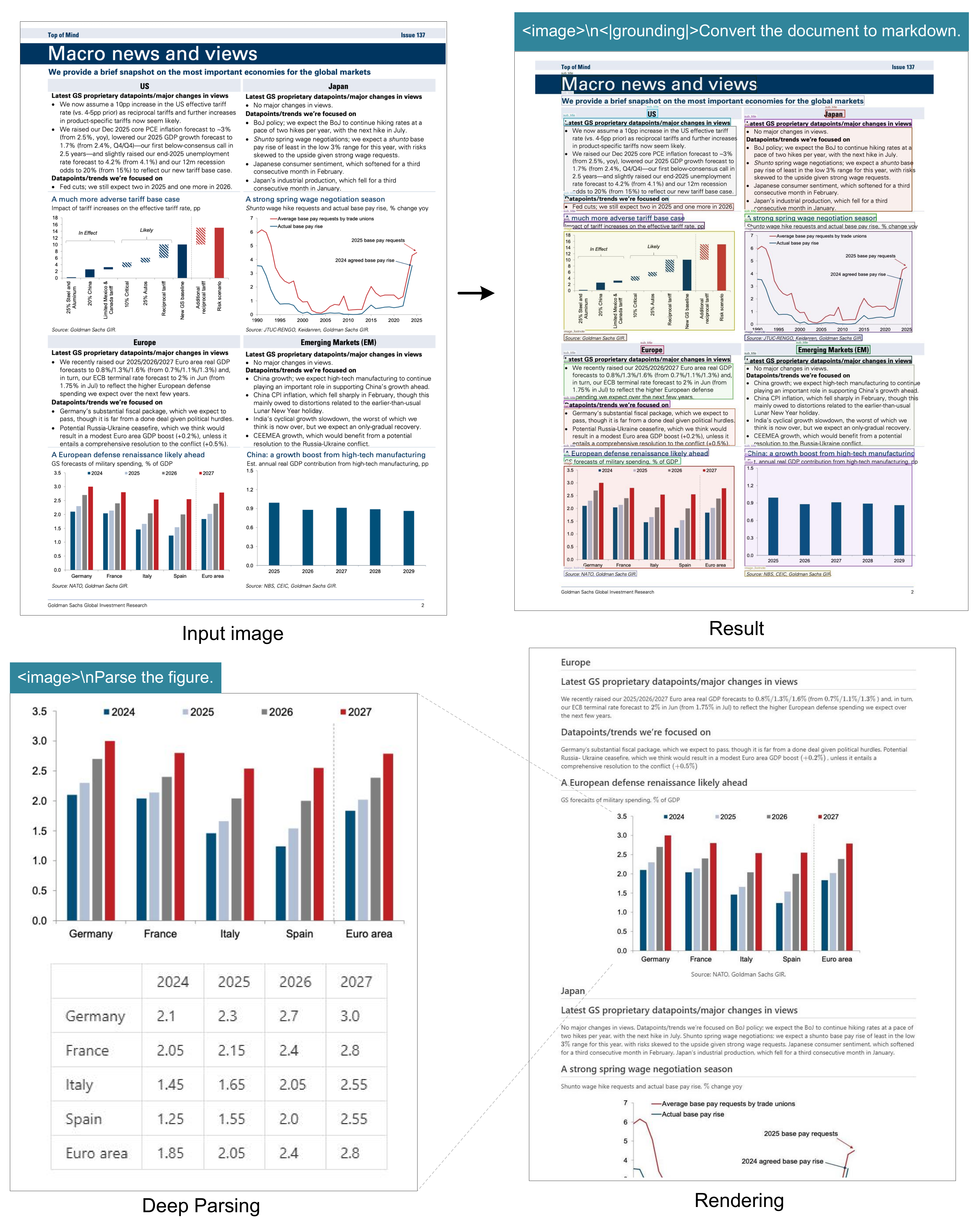}
	\caption{In the field of financial research reports, the deep parsing mode of DeepSeek-OCR can be used to obtain structured  results of charts within documents. Charts are a crucial form of data representation in finance and scientific fields, and the chart structured extraction is an indispensable capability for future OCR models.}
	\label{fig:11}
\end{figure}

\begin{figure}[p]
	\centering
    \includegraphics[width=1.0\linewidth]{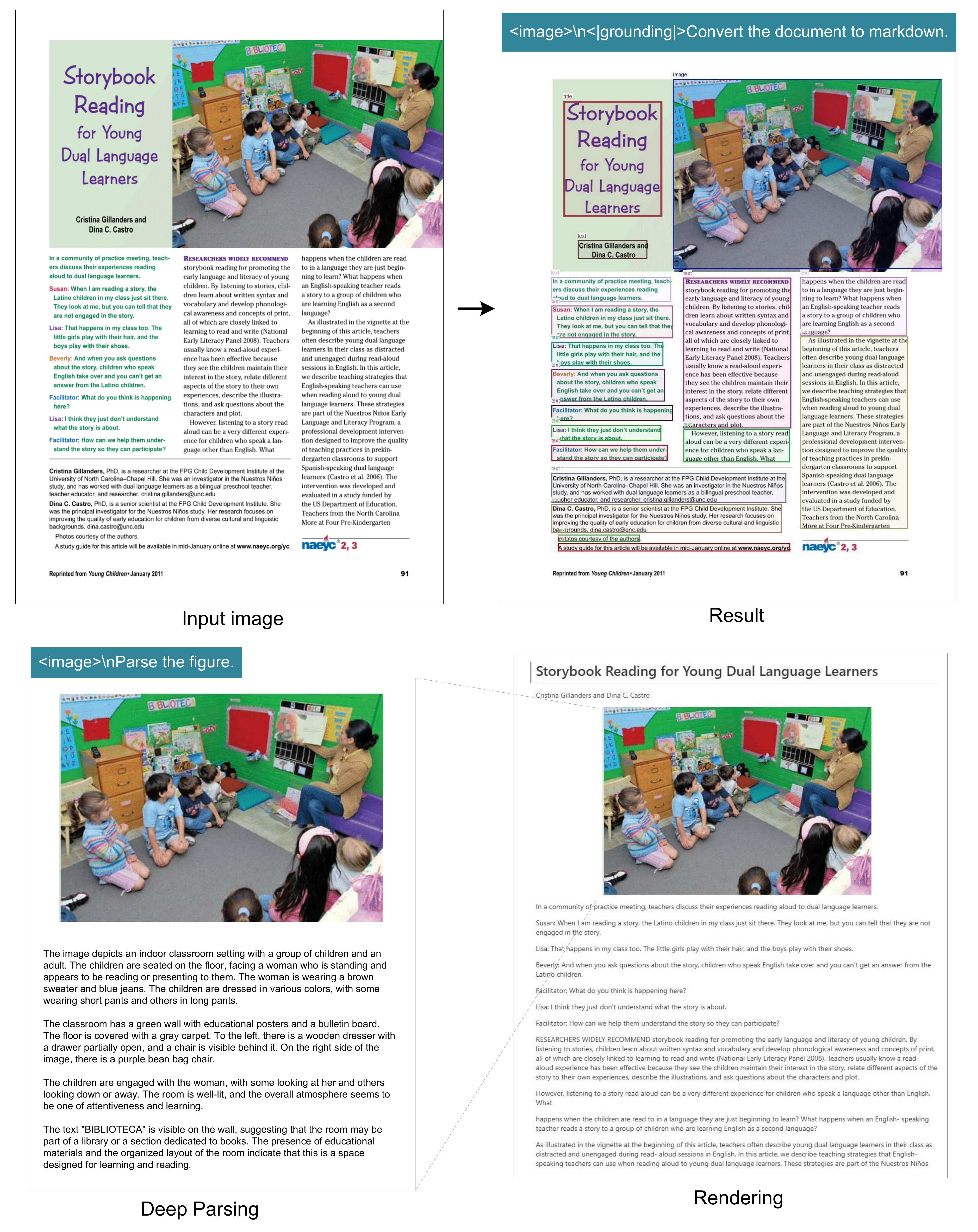}
	\caption{For books and articles, the deep parsing mode can output dense captions for natural images in the documents. With just a prompt, the model can automatically identify what type of image it is and output the required results. }
	\label{fig:12}
\end{figure}

\begin{figure}[p]
	\centering
    \includegraphics[width=1.0\linewidth]{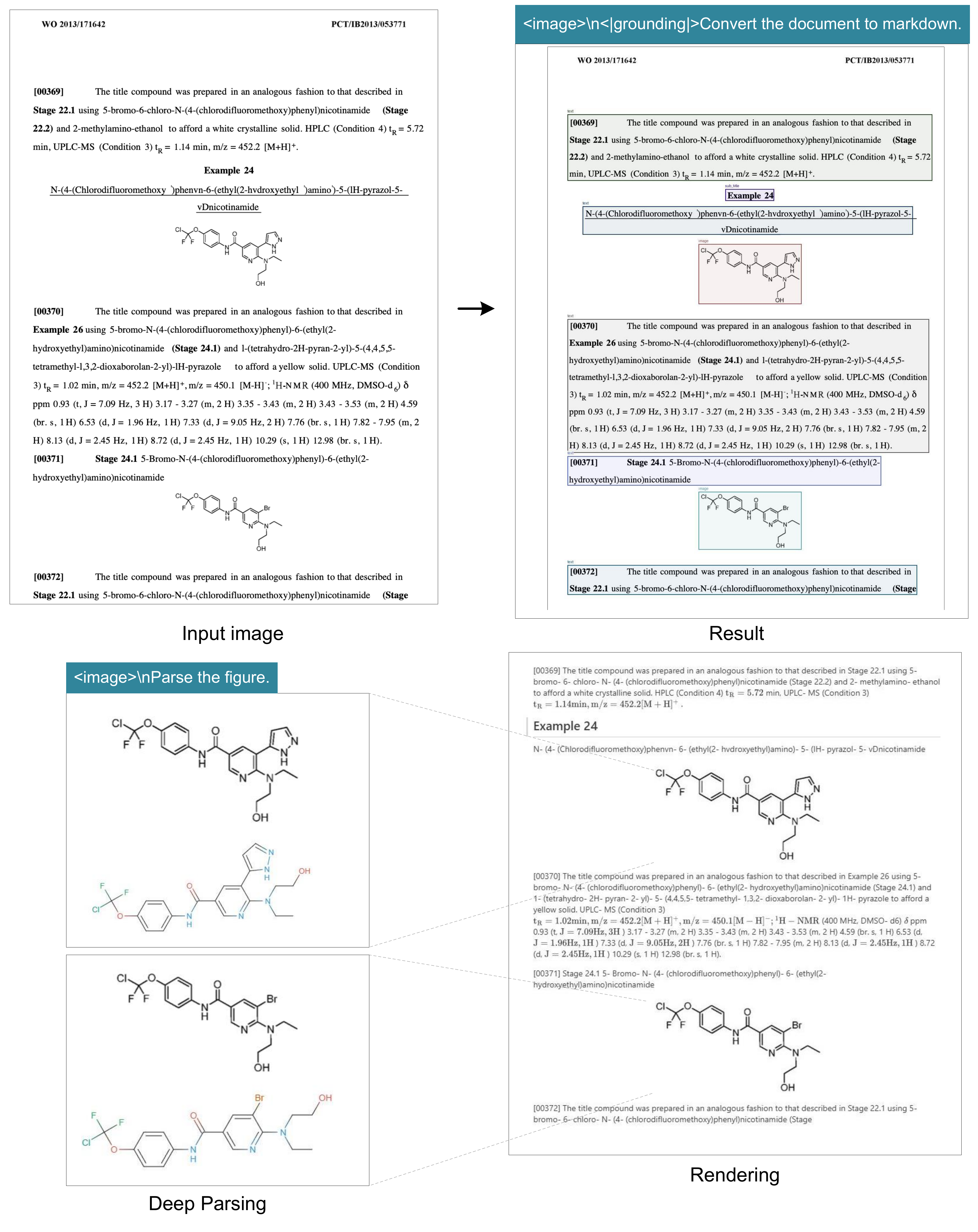}
	\caption{DeepSeek-OCR in deep parsing mode can also recognize chemical formulas within chemical documents and convert them to SMILES format. In the future, OCR 1.0+2.0 technology may play a significant role in the development of VLM/LLM in STEM fields. }
	\label{fig:13}
\end{figure}

\begin{figure}[!h]
	\centering
    \includegraphics[width=1.0\linewidth]{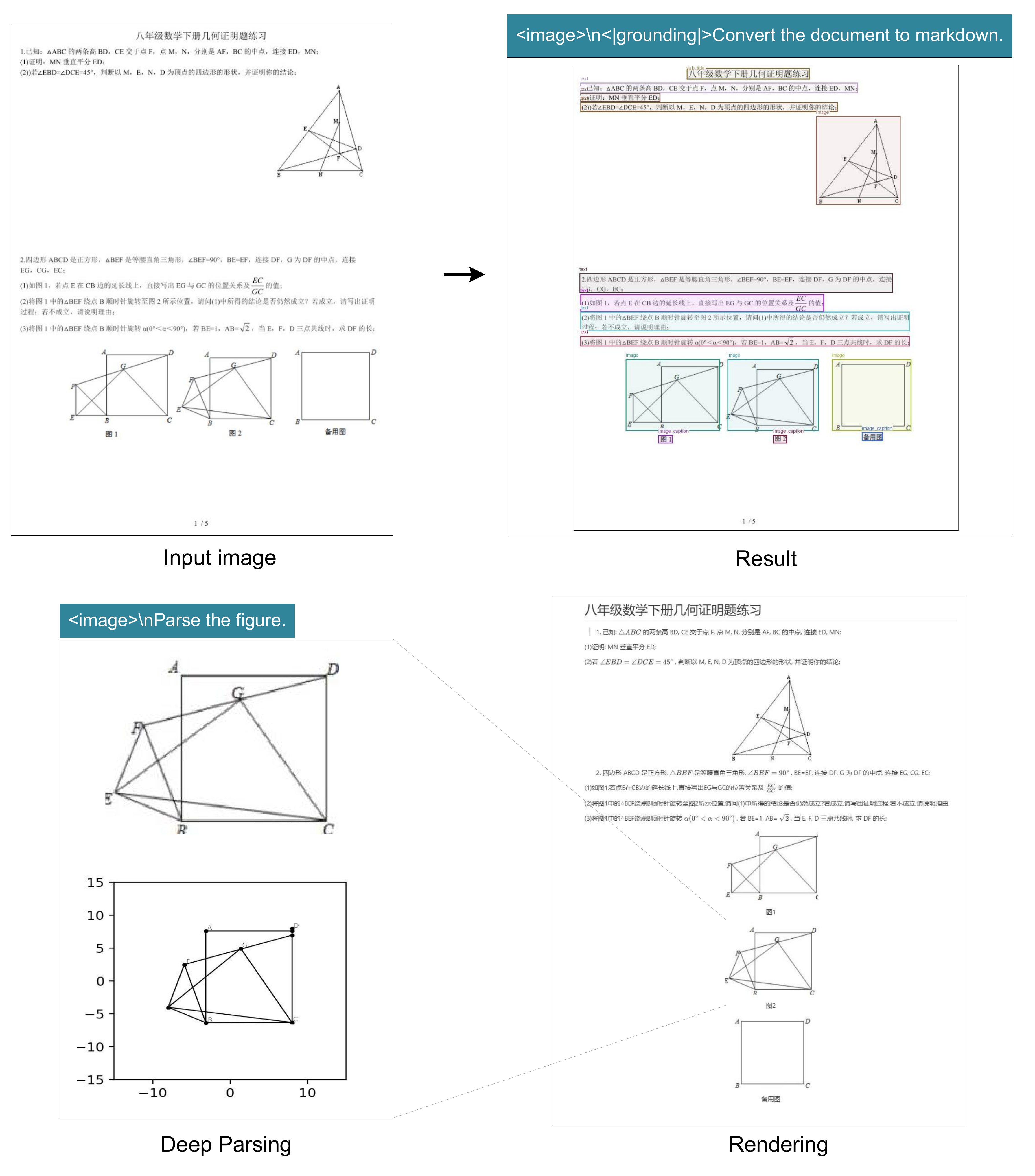}
	\caption{DeepSeek-OCR also possesses the capability to copy (structure) simple planar geometric figures. Due to the intricate interdependencies among line segments in geometric shapes, parsing geometry task is extremely challenging and has a long way to go.}
	\label{fig:14}
\end{figure}

DeepSeek-OCR possesses both layout and OCR 2.0 capabilities, enabling it to further parse images within documents through secondary model calls, a feature we refer to as "deep parsing". As shown in Figures~\ref{fig:11},\ref{fig:12},\ref{fig:13},\ref{fig:14}, our model can perform deep parsing on charts, geometry, chemical formulas, and even natural images, requiring only a unified prompt.

\newpage

\subsubsection{Multilingual recognition}

\begin{figure}[!h]
	\centering
    \includegraphics[width=1.0\linewidth]{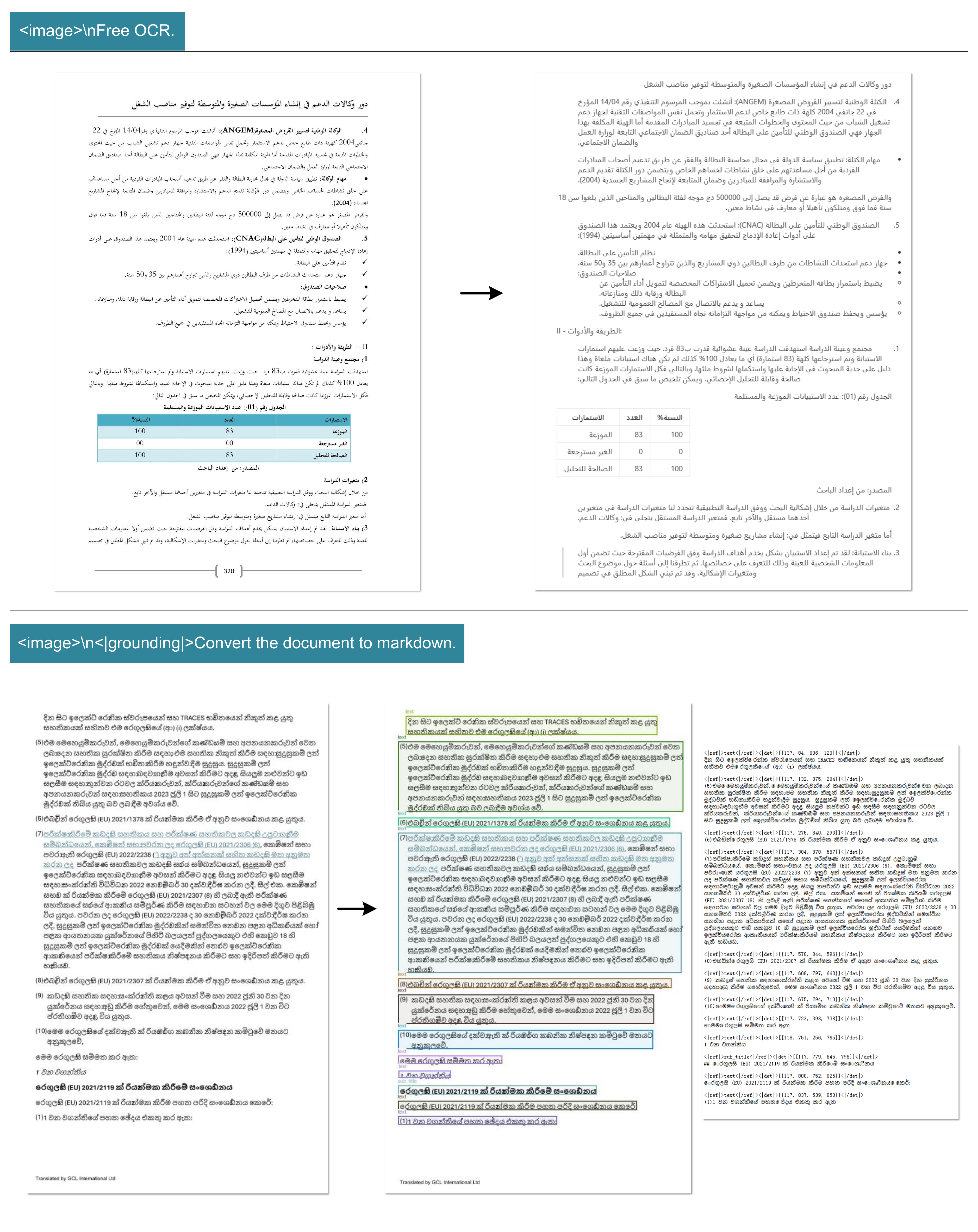}
	\caption{To endow the capability of processing widely crawled PDFs (multilingual data), we train our model with OCR capabilities for nearly 100 languages. Minority language documents can also support both layout and non-layout outputs through different prompts.}
	\label{fig:15}
\end{figure}

PDF data on the Internet contains not only Chinese and English, but also a large amount of multilingual data, which is also crucial when training LLMs. For PDF documents, DeepSeek-OCR can handle nearly 100 languages. Like Chinese and English documents, multilingual data also supports both layout and non-layout OCR formats. The visualization results are shown in Figure~\ref{fig:15}, where we select Arabic and Sinhala languages to demonstrate results.

\subsubsection{General vision understanding}
\begin{figure}[!h]
	\centering
    \includegraphics[width=1.0\linewidth]{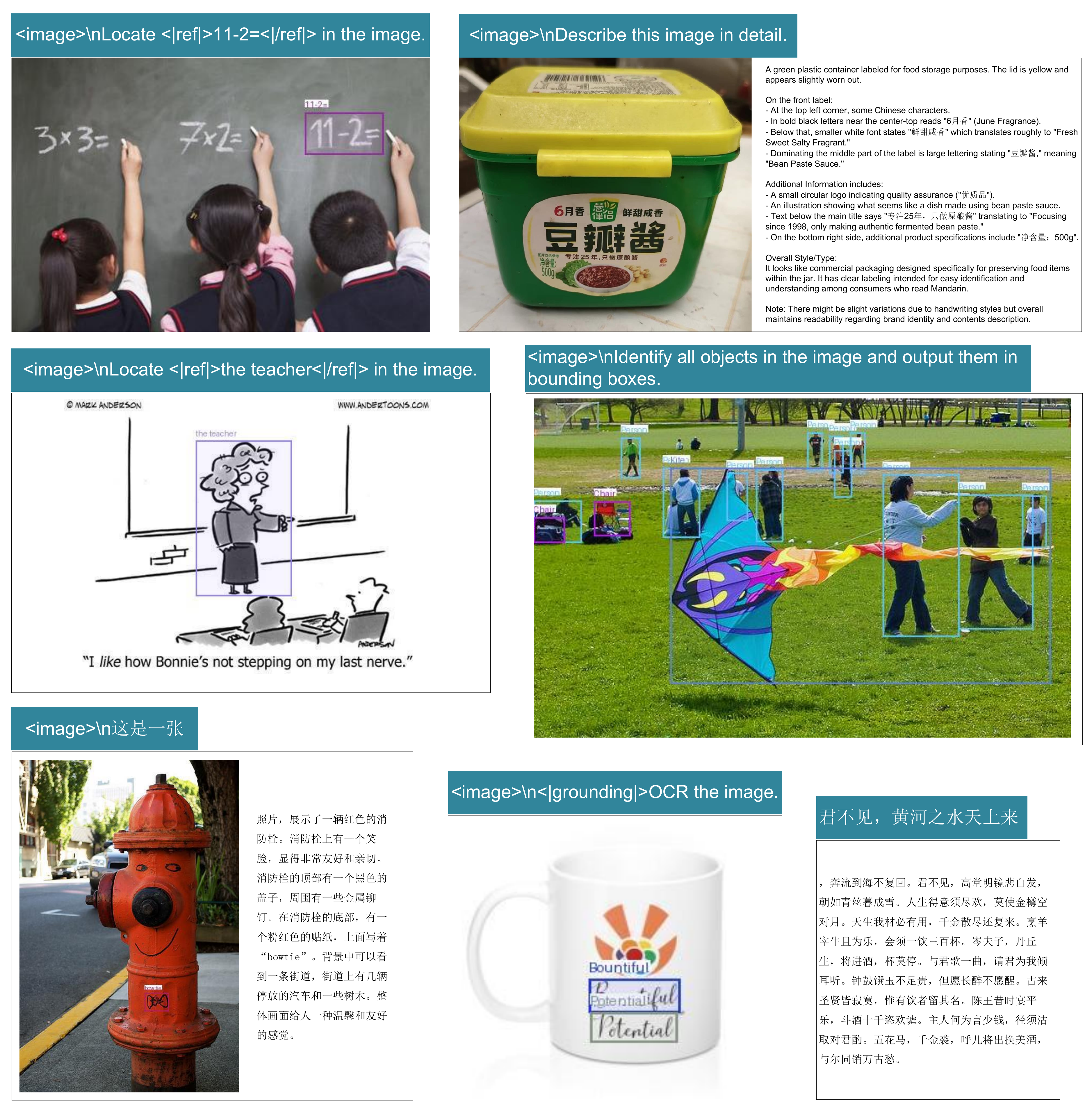}
	\caption{We retain DeepSeek-OCR's capabilities in general visual understanding, mainly including image description, object detection, grounding, etc. Meanwhile, due to the inclusion of text-only data, DeepSeek-OCR's language capabilities are also retained. Note that since we do not include SFT (Supervised Fine-Tuning) stage, the model is not a chatbot, and some capabilities need completion prompts to be activated.}
	\label{fig:16}
\end{figure}

We also provide DeepSeek-OCR with a certain degree of general image understanding capabilities. The related visualization results are shown in Figure~\ref{fig:16}.

\begin{figure}[!t]
	\centering
    \includegraphics[width=1.0\linewidth]{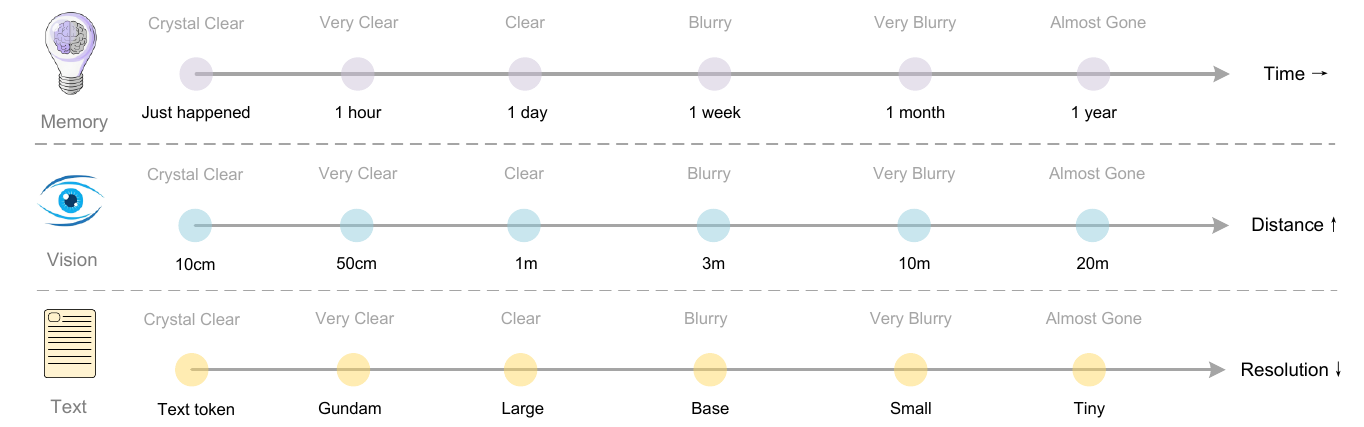}
	\caption{Forgetting mechanisms constitute one of the most fundamental characteristics of human memory. The contexts optical compression approach can simulate this mechanism by rendering previous rounds of historical text onto images for initial compression, then progressively resizing older images to achieve multi-level compression, where token counts gradually decrease and text becomes increasingly blurred, thereby accomplishing textual forgetting.}
	\label{fig:17}
\end{figure}

\section{Discussion}
Our work represents an initial exploration into the boundaries of vision-text compression, investigating how many vision tokens are required to decode $N$ text tokens. The preliminary results are encouraging: DeepSeek-OCR achieves near-lossless OCR compression at approximately 10$\times$ ratios, while 20$\times$ compression still retains 60\% accuracy. These findings suggest promising directions for future applications, such as implementing optical processing for dialogue histories beyond $k$ rounds in multi-turn conversations to achieve 10$\times$ compression efficiency.

For older contexts, we could progressively downsizing the rendered images to further reduce token consumption. This assumption draws inspiration from the natural parallel between human memory decay over time and visual perception degradation over spatial distance—both exhibit similar patterns of progressive information loss, as shown in Figure~\ref{fig:17}. By combining these mechanisms, contexts optical compression method enables a form of memory decay that mirrors biological forgetting curves, where recent information maintains high fidelity while distant memories naturally fade through increased compression ratios.

While our initial exploration shows potential for scalable ultra-long context processing, where recent contexts preserve high resolution and older contexts consume fewer resources, we acknowledge this is early-stage work that requires further investigation. The approach suggests a path toward theoretically unlimited context architectures that balance information retention with computational constraints, though the practical implications and limitations of such vision-text compression systems warrant deeper study in future research.

\section{Conclusion}
In this technical report, we propose DeepSeek-OCR and preliminarily validate the feasibility of contexts optical compression through this model, demonstrating that the model can effectively decode text tokens exceeding 10 times the quantity from a small number of vision tokens. We believe this finding will facilitate the development of VLMs and LLMs in the future. Additionally, DeepSeek-OCR is a highly practical model capable of large-scale pretraining data production, serving as an indispensable assistant for LLMs. Of course, OCR alone is insufficient to fully validate true context optical compression and we will conduct digital-optical text interleaved pretraining, needle-in-a-haystack testing, and other evaluations in the future. From another perspective, optical contexts compression still offers substantial room for research and improvement, representing a promising new direction.

\newpage

\bibliography{main}

\end{CJK*}
\end{document}